\documentclass[10pt,twocolumn,letterpaper]{article}

\usepackage{cvpr}              %

\definecolor{iccvblue}{rgb}{0.21,0.49,0.74}
\usepackage[pagebackref,breaklinks,colorlinks]{hyperref}
\hypersetup{
colorlinks=true,
allcolors=iccvblue,
}

\usepackage{graphicx}
\usepackage{amsmath}
\usepackage{amssymb}
\usepackage{booktabs}
\usepackage{algorithm}
\usepackage{algpseudocode}
\usepackage[dvipsnames]{xcolor}
\usepackage[normalem]{ulem}
\usepackage{scalerel,graphicx,xparse}
\usepackage{sidecap}
\sidecaptionvpos{figure}{t}
\usepackage{soul}

\newtheorem{problem}{Problem}

\usepackage[capitalize,nameinlink]{cleveref}
\crefdefaultlabelformat{#2\textbf{#1}#3}
\crefname{equation}{}{}
\creflabelformat{equation}{#2\textup{\bf #1}#3}
\crefname{equation}{Eq.}{Eqs.}
\Crefname{equation}{Equation}{Equations}
\crefname{figure}{Fig.}{Figs.}
\crefname{subfigure}{Fig.}{Figs.}
\Crefname{figure}{Figure}{Figures}
\Crefname{subfigure}{Figure}{Figures}
\crefname{table}{Tab.}{Tabs.}
\Crefname{table}{Table}{Tables}
\crefname{section}{Sec.}{Secs.}
\Crefname{section}{Section}{Sections}
\crefname{problem}{Problem}{Problems}
\Crefname{problem}{Problem}{Problems}
\crefname{definition}{Definition}{Definitions}
\Crefname{definition}{Definition}{Definitions}
\crefname{lemma}{Lemma}{Lemmas}
\Crefname{lemma}{Lemma}{Lemmas}
\crefname{theorem}{Thm.}{Thms.}
\Crefname{theorem}{Theorem}{Theorems}
\crefname{remark}{Rmk.}{Rmks.}
\Crefname{remark}{Remark}{Remarks}
\crefname{enumi}{item}{items}
\Crefname{enumi}{Item}{Items}
\crefname{algocf}{Alg.}{Algs.}
\Crefname{algocf}{Algorithm}{Algorithms}
\crefname{assumption}{Asm.}{Asms.}
\Crefname{assumption}{Assumption}{Assumptions}
\crefname{ALC@unique}{line}{lines}
\Crefname{ALC@unique}{Line}{Lines}
\usepackage{dsfont}

\definecolor{vividp}{HTML}{FF6F00}

\newcommand{\del}[1]{{\color{lightgray}{}}}

\newcommand{\vect}[1]{\mathbf{#1}}
\newcommand{\vectnorm}[1]{\overline{\mathbf{#1}}}

\newcommand{\rotp}[1]{\tilde{\mathbf{#1}}}

\newcommand{\ds}[1]{#1^{\circ}}

\newcommand{\uf}[1]{\underline{#1}}
\usepackage{nicematrix}
\usepackage{booktabs}
\usepackage{makecell}
\usepackage{multirow}
\usepackage{colortbl}

\def\paperID{46634} %
\def\confName[final]{CVPR}
\def\confYear{2026}

\title{FLIGHT: \underline{F}ibonacci \underline{L}attice-based \underline{I}nference for \underline{G}eometric \underline{H}eading in real-\underline{T}ime}

\author{
David Dirnfeld$^1$ \and
Fabien Delattre$^1$ \and
Pedro Miraldo$^2$ \and
Erik Learned-Miller$^1$  \and \\
$^1$University of Massachusetts Amherst \ \ \
$^2$Mitsubishi Electric Research Laboratories (MERL)
\ \ \
}

\begin{document}
\maketitle

\begin{abstract}
Estimating camera motion from monocular video is a fundamental problem in computer vision, central to tasks such as SLAM, visual odometry, and structure-from-motion. Existing methods that recover the camera’s heading under known rotation, whether from an IMU or an optimization algorithm, tend to perform well in low-noise, low-outlier conditions, but often decrease in accuracy or become computationally expensive as noise and outlier levels increase. To address these limitations, we propose a novel generalization of the Hough transform on the unit sphere ($\mathcal{S}^2$) to estimate the camera's heading. First, the method extracts correspondences between two frames and generates a great circle of directions compatible with each pair of correspondences. Then, by discretizing the unit sphere using a Fibonacci lattice as bin centers, each great circle casts votes for a range of directions, ensuring that features unaffected by noise or dynamic objects vote consistently for the correct motion direction. Experimental results on three datasets demonstrate that the proposed method is on the Pareto frontier of accuracy versus efficiency. Additionally, experiments on SLAM show that the proposed method reduces RMSE by correcting the heading during camera pose initialization.
\end{abstract}

\section{Introduction}\label{sec:intro}
\begin{figure}[t]
  \centering
     \includegraphics[width=1\linewidth]{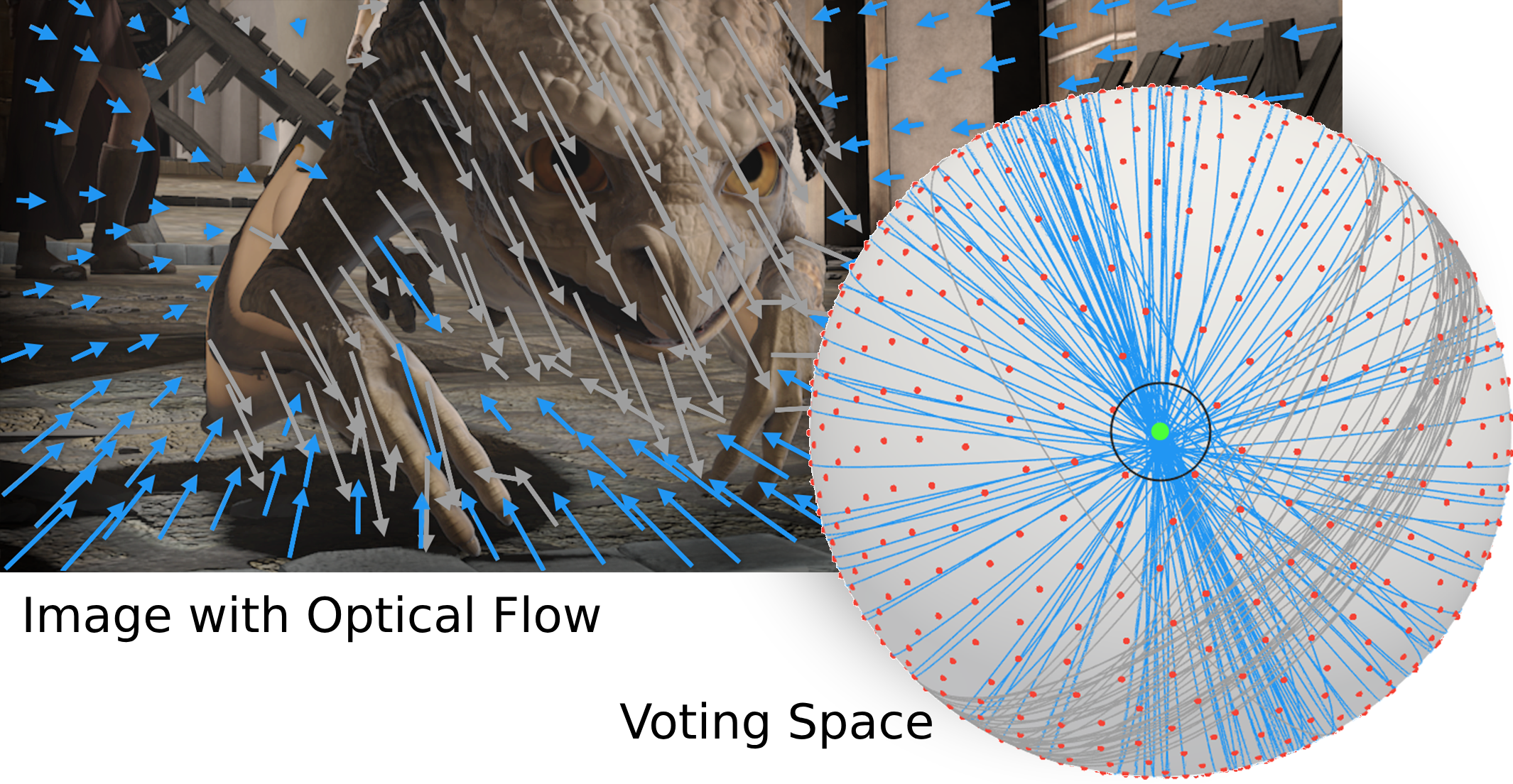}
    \caption{{\bf Left.} 
    A frame of a dynamic scene from the Sintel dataset. The blue vectors show translational flows compatible with the winning Fibonacci bin $\vect{s}_j$ as shown by the green point. Gray vectors show translational flows that do not agree on a direction. {\bf Right.} The unit sphere of the possible directions a camera can move. The red points show a sparse sampling of the unit sphere (i.e.,~Fibonacci bins) with our hierarchical approach. The blue great circles (corresponding to the blue flow vectors in the frame) represent great circles that vote for a unique direction, whereas the gray great circles do not.%
    }
   \label{fig:cover}
\end{figure}
Imagine a movie whose opening scene is shot by a camera moving through a crowded marketplace. As long as the camera motion is smooth (rather than extremely jerky), we typically have no trouble inferring the camera's motion. We instinctively interpret this view as if we're following a character moving forward, avoiding obstacles, stepping in open areas, and so on. We feel as if we are there, experiencing the motion as we watch the video. The ability to infer complex motion from a video with moving objects in a geometrically rich scene is practically effortless for humans. Yet, computer vision algorithms for understanding motion in monocular video still face significant challenges.

The standard algorithmic approach for pose estimation is to jointly estimate camera rotation and translation \cite{Nistr2004AnES, Fraundorfer2010AMC, 590022, young19883, 211464}. However, in many scenarios, camera rotation and translation can be estimated independently. This work assumes that the camera experiences only translational motion or that the rotation parameters are known. This can occur, for example, when: 1) an inertial measurement unit (IMU) is available; 2) the rotation has been estimated independently from the translation, e.g., \cite{Delattre2023RobustRotation}; or 3) the rotation is fixed as part of an alternating minimization method, e.g., \cite{Campos2019POSEAMMAU}. Focusing only on the heading (we use heading and direction interchangeably) eliminates unnecessary computations in methods that estimate rotation and translation (5DoF). Note: we focus on heading estimation, since without additional information (such as the size of a calibration device), it is, in general, impossible to determine the {\em magnitude} of a translation since the same pair of video frames could, in principle, be created from scenes with different scales. This is the speed-distance ambiguity.

One of the main challenges in translation heading estimation, and in camera motion estimation more broadly, is developing approaches that remain robust in the presence of noise and outliers. Such perturbations arise from inaccurate feature estimations, erroneous correspondences, or moving objects. Several works \cite{LonguetHiggins1980TheIO, Guissin1991DirectCO, 232074, NEGAHDARIPOUR1989303, horn1987computationally} forgo this problem and operate under the assumption of minimal noise and outliers. Alternatively, other methods explicitly address these issues \cite{Fredriksson2014FastAR, Fredriksson2015PracticalRT, Heikkonen1995Recovering3M} and handle outliers by implementing a sampling strategy. However, these sampling strategies increase the computational cost as the number of outliers increases, particularly in complex scenes.

In this paper, we propose FLIGHT, a voting technique for camera translation that uses a novel generalization of the Hough transform.  Unlike \cite{Fredriksson2014FastAR, Fredriksson2015PracticalRT, Heikkonen1995Recovering3M}, our sampling strategy does not drastically increase the computational cost. Instead, we generate translation heading hypotheses compatible with each feature correspondence. We then adopt a voting scheme to estimate the translation heading as shown in \cref{fig:cover}. We show improvements over four existing works in terms of time and accuracy, which are valuable in applications such as robotics, drone navigation, and other real-time applications. In addition, improvements in these techniques open up new avenues for improving SLAM \cite{MurArtal2015ORBSLAMAV, Qin2017VINSMonoAR, Tateno2017CNNSLAMRD, freda2025pyslamopensourcemodularextensible}, visual odometry \cite{Nistr2004VisualO,Wang_2017}, structure from motion (SfM) \cite{Schnberger2016StructurefromMotionR, Mildenhall2020NeRF, Cui2017HSfMHS, Zhu2023NICERSLAMNI}, and many other applications.

\del{We focus on direction estimation since without using additional sources of information (such as the size of a calibration device), it is, in general, impossible to determine the magnitude of translation since the same pair of video frames could, in principle, be created from scenes with different scales. This is the speed-distance ambiguity. For this reason, we concentrate only on estimating the direction of motion and not the magnitude. }

To summarize, our contributions to direction estimation are as follows:
\begin{enumerate}
    \item We propose a novel solution for estimating the direction of motion from monocular video using a generalized Hough transform on the unit sphere;
    \item We implement a hierarchical discretization approach to improve efficiency;
    \item We introduce a non-linear refinement on the unit-sphere to improve the direction estimate further;
    \item We show FLIGHT improves the SLAM pipeline when used in the camera pose initialization step.  
\end{enumerate}

\section{Related work}

\paragraph{2-Point Translation Estimation:} In \cite{Sun20002pointLA}, the authors propose a method that aggregates the coefficients from two-point correspondences into a matrix and estimates the translation using an SVD approach. In contrast, \cite{6907672} presents an alternative method whereby image features are first projected onto the unit sphere, impose the epipolar constraint and solve for the translation using a two-point RANSAC algorithm. In \cite{Chou20152pointRF}, the authors implement a two-stage approach wherein initial motion estimation is estimated using the epipolar constraint with two-point correspondences, followed by outlier rejection through homography matching. While primarily focused on image matching, the underlying principles can be adapted for camera heading estimation. Over the years, a multitude of sampling strategies have emerged, including \cite{Chum2005MatchingWP, Barth2019ProgressiveNS, Chum2003LocallyOR, barath2019magsacplusplus, piedade2023bansacdynamicbayesiannetwork}. Yet, a significant challenge with RANSAC-based methods is their computational complexity, which is on the order of $\mathcal{O}(n^3)$.

\paragraph{Unit Sphere Translation Direction Estimation:} 
In \cite{Enqvist}, the authors solve for translation using a linear programming scheme under the assumption that no outliers are present in the data. To estimate the translation when there are outliers, \cite{Fredriksson2014FastAR}, present a provably optimal algorithm that, given an error threshold $\epsilon$, finds the translation direction that maximizes the number of correspondences with reprojection errors below $\epsilon$. The time complexity of the proposed algorithm is $\mathcal{O}(n^2\log(n))$. In \cite{Fredriksson2015PracticalRT}, the authors propose a branch-and-bound approach to speed up \cite{Fredriksson2014FastAR} while keeping the same kind of theoretical guarantees. The authors divide the parameter space $S^2$ using spherical triangles. Unlike \cite{Fredriksson2014FastAR, Fredriksson2015PracticalRT}, our method does not generate hypotheses from pairs of correspondences. Instead, it considers the entire set of possible solutions for each correspondence. It has a time complexity of $\mathcal{O}(nm)$ where $n$ is the number of correspondences and $m$ is the number of bins in the histogram on $S^2$. 
\paragraph{Focus of Expansion (FOE) Estimation:} A closely related family of methods estimates motion direction by finding the Focus of Expansion (FOE), the point at which the 2D motion field becomes zero \cite{LonguetHiggins1980TheIO, burger1990estimating, 211464, 211466, horn1987computationally, Guissin1991DirectCO}. We consider two key baseline approaches. In \cite{PassiveNU}, the authors derive the FOE by solving a system of linear equations in a least-squares form given pairs of correspondences. By back-projecting the FOE, the solution is an estimate of the heading direction. However, this method is inherently sensitive to outliers because it treats all feature pairs equally. To address this, \cite{Heikkonen1995Recovering3M} introduces a randomized Hough transform algorithm. Their method builds an accumulator space in spherical coordinates. First, two random feature pairs are selected to compute a candidate FOE, which is then projected onto the unit sphere. The candidate is assigned to an existing cell in the accumulator if it falls within a predefined boundary, or a new cell is created otherwise. Although this method is conceptually similar to our approach, its runtime complexity is on the order of $\mathcal{O}(n^2m)$, considering there are $n^2$ pairs and $m$ accumulator cells whereas FLIGHT $\mathcal{O}(nm)$, and our experiments (discussed in \cref{sec:experiments}) also demonstrate superior accuracy.

\section{Problem Definition}\label{sec:background}

\del{\subsection{Notation}
In this work, we use regular non-bold letters for scalars, e.g., $a\in \mathbb{R}$, and vectors by small bold letters, such as $\mathbf{a}\in \mathbb{R}^d = [a_1, \dots, a_d]$. To distinguish flow vectors from other vectors, we use the arrow, such as $\vec{\mathbf{a}} \in \mathbb{R}^d$. For any vector of dimension $d$, $\mathbf{a}\in\mathbb{R}^d$, its direction is given by $\overline{\mathbf{a}} \in S^{d-1} = \tfrac{\mathbf{a}}{\|\mathbf{a}\|}$. We represent matrices by bold capital letters, such as $\mathbf{A}\in \mathbb{R}^{d\times m}$.  We use calligraphy letters for representing sets, such as $\mathcal{A} = \{\mathbf{a}_1, \cdots, \mathbf{a}_n\}$, for a set of vectors, but also applies to scalars and flow vectors. Finally, to distinguish between features in the 3D and image coordinates, we use superscripts, as $\mathbf{p}^{W} \in \mathbb{R}^3$ and $\mathbf{p} \in \mathbb{R}^2$ for world and image points, respectively.}

Let $\textbf{p}^{W}$ be a point in the world, and $\textbf{p}, \textbf{q}$ be the normalized homogeneous projections of $\textbf{p}^{W}$ in the first and second frames respectively. From epipolar geometry, we know that $\textbf{p}$ and $\textbf{q}$ are related by the essential matrix $\textbf{E}\in\mathbb{R}^{3 \times 3}$:
\begin{align}\label{eq:essential}
    \textbf{q}\textbf{E}\textbf{p} = 0, \ \ \text{with} \
    \textbf{E} = [\textbf{t}]_\times \textbf{R}
\end{align}
where $\textbf{t}\in\mathbb{R}^3$ is the translation and $\textbf{R}\in\mathcal{SO}(3)$ is the camera rotation, and $[\cdot]_\times$ is the matrix representation of the cross product, such as $\mathbf{a}\times \mathbf{b} = [\mathbf{a}]_\times \mathbf{b}$. Notice that \cref{eq:essential} holds for any scaled translation $\lambda \mathbf{t}$ with $\lambda \in \mathbb{R}$, reflecting the fact that relative pose estimation recovers only five of the six degrees of freedom: the rotation $\mathbf{R}$ and the direction of translation $\vectnorm{t} = \tfrac{\mathbf{t}}{|\mathbf{t}|}$, but not its magnitude.

Since we assume $\textbf{R}$ is known and doesn't need to be estimated, we define the rotation-compensated correspondences $\rotp{p} = \textbf{R}\textbf{p}$ as the point $\textbf{p}$ rotated into $\textbf{q}$ frame of reference, and rewrite \cref{eq:essential} and solve for translation \cite{Kneip2012} as 
\begin{equation}
\textbf{q}\cdot\left(\textbf{t} \times \rotp{p}\right) = 0 \ \ \equiv \ 
\textbf{t}\cdot\left(\rotp{p} \times \textbf{q}\right) = 0,
\label{eq:translation_set}
\end{equation}
which is used to define the problem we solve below.

\begin{problem}
\label{problem}
For a given set of matches $\{ \mathbf{p} \leftrightarrow \mathbf{q} \}$ and a rotation $\mathbf{R}$, we aim to find the translation direction (camera heading) defined as $\vectnorm{t} \in S^2$ satisfying the constraint in \cref{eq:translation_set} for the largest consensus set of the matches.
\end{problem}

\section{FLIGHT}\label{sec:method}

In this work, we aim to find the camera heading, $\vectnorm{t}$, that satisfies the problem outlined in \cref{problem}, given a set of image correspondences $\{(\vect{p}_1,\vect{q}_1) \dots, (\vect{p}_N,\vect{q}_N)\}$.
These correspondences can be obtained either through discrete feature detection and matching algorithms (e.g., \cite{surf, Lowe1999ObjectRF, DeTone2017SuperPointSI}) or by computing a dense optical flow field \cite{rapidflow, gmaflow, flownets, csflow}. In either case, we remove the rotational component from $\textbf{p}_i$, as has been done \cite{Pia-itsmoving, Bideau_2018_CVPR, Bideau2022TheRS}, to obtain $\{(\rotp{p}_1,\vect{q}_1) \dots, (\rotp{p}_N,\vect{q}_N)\}$. The resulting correspondences are, therefore, only influenced by the camera translation and possibly noise and/or moving objects.

\subsection{Great circle of compatible motions}\label{sec:compatible_motions}
\begin{figure}
    \centering
    \includegraphics[width=.9\linewidth]{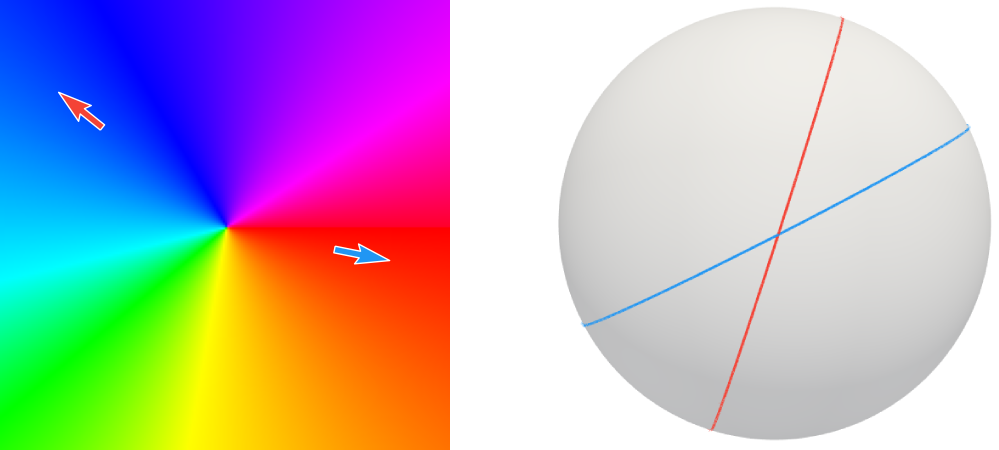}
    \caption{In the case of pure translation flow, two pairs of point correspondences are needed to obtain the direction of motion. On the left, we sample two flow vectors (indicated by the red and blue arrows). On the right, we show the great circles of translation directions compatible with these flow vectors. The points of intersection are the only directions compatible with both flow vectors.}
    \label{fig:solution_from_two_great_circle}
\end{figure}
Given a pair of rotation-compensated correspondences $\rotp{p}_i$ and $\textbf{q}_i$, the set of compatible translations that map $\{\rotp{p}_i \leftrightarrow\textbf{q}_i\}$ lies on a plane $\mathcal{P}_i$ \cite{Fredriksson2015PracticalRT} that passes through the origin and has a normal as defined according to \cref{eq:translation_set}, namely
\begin{align}\label{eq:gc_normal}
    \textbf{n}_i = \rotp{p}_i \times \textbf{q}_i, 
\end{align}
with $\textbf{n}_i$ representing the normal vector to the plane $\mathcal{P}_i$.

Since our focus is solely on the set of compatible translation directions, that is, the set of \emph{normalized} translation vectors, we only consider the intersection between the plane $\mathcal{P}_i$ and the unit sphere $S^2$. This intersection yields a great circle on $S^2$ with normal vector $\textbf{n}_i$.

\begin{figure}[t]
     \centering
     \begin{subfigure}[b]{0.2\textwidth}
         \centering
         \includegraphics[width=\textwidth]{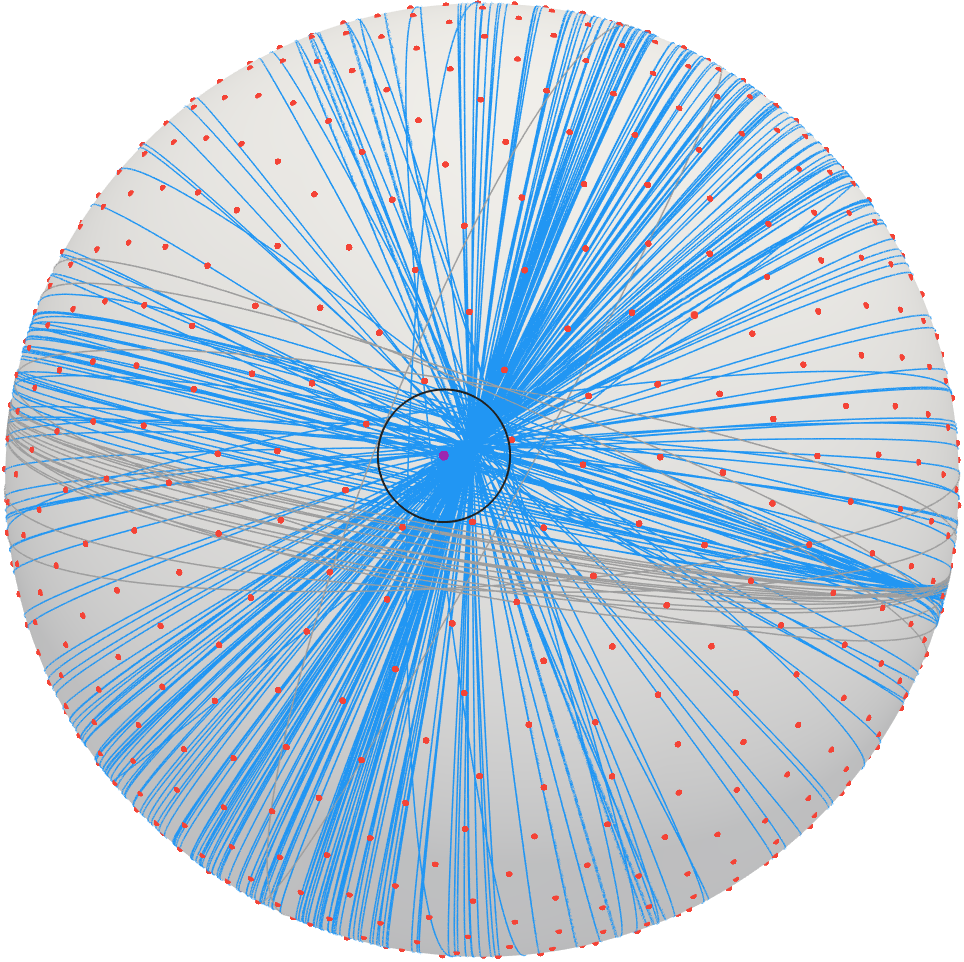}
         \label{fig:sparse_fib_shpere}
     \end{subfigure}
     \hfill
     \begin{subfigure}[b]{0.2\textwidth}
         \centering
         \includegraphics[width=\textwidth]{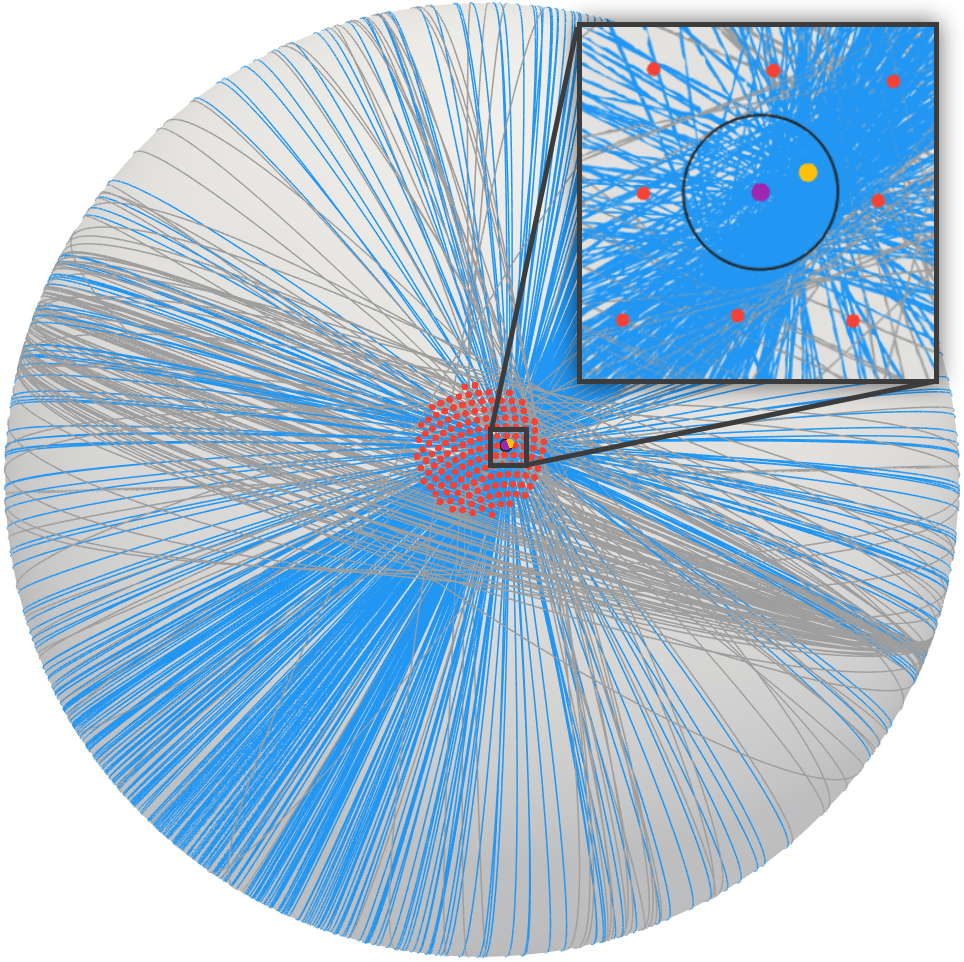}
         \label{fig:dense_fib_shpere}
     \end{subfigure}
        \caption{We employ a 2-stage voting scheme on the unit sphere $\mathcal{S}^2$ representation of the possible camera translation directions between two consecutive video frames using a {\em Fibonacci lattice}. \textbf{Left:} In the first stage, the red points represent a sparse Fibonacci lattice. The blue and gray arcs across the sphere are the {\em great circles} representing the camera motion directions compatible with a {\bf single pair of feature correspondences}. Our goal is to find the point on the sphere compatible with the largest number of features. \textbf{Right:} In the second stage, the red points represent a dense sampling of the Fibonacci lattice in the winning region of step 1. The yellow dot shows the ground truth translation, and the purple point is the winning bin.}
        \label{fig:hierarchical_approach}
\end{figure}

\subsection{Voting on the unit sphere}\label{sec:voting}
\begin{figure}
    \centering
    \begin{subfigure}{.2\textwidth}
      \centering
      \includegraphics[width=.9\linewidth]{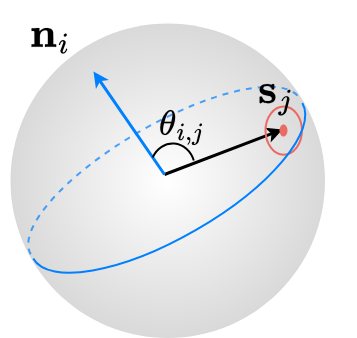}
      \caption{High-level view.}
      \label{fig:sub1}
    \end{subfigure}%
    \begin{subfigure}{.2\textwidth}
      \centering
      \includegraphics[width=.9\linewidth]{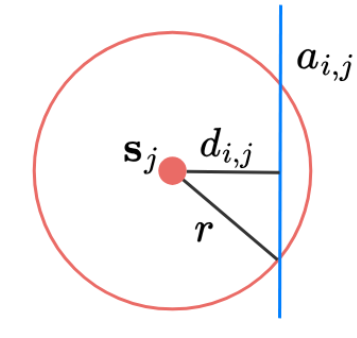}
      \caption{The bin at $\vect{s}_j$.}
      \label{fig:sub2}
    \end{subfigure}
    \caption{The relationship of the angle $\theta_{i, j}$ between the normal $\vect{n}_i$ of a great circle $i$ and a Fibonacci bin center $\vect{s}_j$. When $\theta_{i, j}$ is close to $\pi/2$, the great circle intersects the region $\vect{s}_j$ defined by $r$. Given the bin is approximately flat, we use the Pythagorean Theorem to solve for $b_{i, j}$. The length $a_{i, j}$ of the intersection is therefore $2b_{i, j}$ and the amount the great circle $i$ votes for $\vect{s}_j$.}
    \label{fig:fib_point_great_circle}
\end{figure}
In the case of pure translation, estimating the motion direction requires two pairs of point correspondences, where each pair defines a great circle of feasible translation directions, as outlined in the preceding subsection. The points where the two great circles intersect are the only solutions compatible with both correspondence pairs. This relationship is illustrated in \cref{fig:solution_from_two_great_circle} for optical flow inputs. In practice, the observed correspondences are affected by moving objects and errors in the match estimation process. To address these effects, we introduce a generalization of the Hough transform voting scheme in which each great circle votes for a set of bins on $S^2$ (the unit sphere). The bin receiving the highest accumulated vote is taken as the estimated motion direction, as illustrated in \cref{fig:hierarchical_approach}. 

\paragraph{Discretizing the unit sphere:} For the Hough transform to remain unbiased, the discretization of $S^2$ should consist of bins that are both uniformly distributed and approximately equal in area. A common strategy is to construct semiregular rectangular cells in spherical coordinates $(\theta, \phi)$ via uniform quantization in $\phi$ and a nonuniform quantization in $\theta$, as in \cite{vanishing_points}. Although this yields bins of roughly equal size, the spacing between bin centers varies substantially over the sphere. To avoid this inconsistency, we place bin centers using a Fibonacci lattice, which yields a near-uniform distribution of points on $S^2$ \cite{alvaro:2010, lin2022deepvanishingpointdetection, HoughParameterSpaceRegularisation}. Let $\mathcal{S} = \{ \vect{s}_1, \dots, \vect{s}_M \}$ denote a Fibonacci lattice comprising of $M$ points. For each $\vect{s}_j \in \mathcal{S}$, we define a bin by a circular region tangent to the sphere centered at $\vect{s}_j$ with a fixed radius $r$. Importantly, the radius $r$ must be large enough to ensure no uncovered regions (i.e., “holes”) on the sphere, which requires overlap between neighboring bins (we discuss how to set $r$ in \cref{subsec:setting_r}). From now on, we will identify each bin solely by its center point $\vect{s}_j$.

\paragraph{Finding the intersections:}  Let $\vect{n}_i$ be the normal to a great circle $i$. The great circle is considered to cast a vote for $\vect{s}_j$ as a candidate for the direction if the great circle intersects the bin $\vect{s}_j$. Mathematically, the angle
\begin{equation}\label{eq:finding_theta}
    \theta_{i, j} = \arccos( | \vect{n}_i\cdot \vect{s}_j |),
\end{equation}
determines the distance (on the surface of the sphere) between the great circle's normal and the bin center. The absolute value appears because $\vect{n}_i$ and $-\vect{n}_i$ represent the same great circle. The angular distance between the great circle $i$ and the center of bin $\vect{s}_j$ is
\begin{equation}
    d_{i, j} = \frac{\pi}{2} - \theta_{i, j}. 
\end{equation}
\Cref{fig:fib_point_great_circle} illustrates the relationship between $\theta_{i, j}$ and $d_{i, j}$.

\paragraph{Weighting the votes:} We want to give higher weight to great circles that have a larger intersection with a bin than those that have smaller intersections. To do this, 
we weight each 
vote by the arc length of the intersection between the great circle and the bin. We define the weight as
\begin{equation}\label{eq:lookup}
    a_{i, j}  = \begin{cases}
                    2b_{i, j} & d_{i, j} < r,\\
                    0 & \text{otherwise,}
                \end{cases}
\end{equation}
where $b_{i,j}=\sqrt{r^2-d^2_{i,j}}$ is half the length of the chord across the bin (see \cref{fig:fib_point_great_circle}).

\paragraph{Winning bin:} Each feature pair defines a great circle with the normal $\vect{n}_i$ as defined in \cref{eq:gc_normal} for which we compute the arc length to each Fibonacci bin $\vect{s}_j$. Let $\mathcal{G} = \{g_1, \cdots, g_M\}$ be the set of the sum of the contributions of each great circle to each Fibonacci bin, where
\begin{equation}\label{eq:sum_of_arcLength}
    g_j = \sum_{i=1}^N a_{i, j}.
\end{equation}
The most likely direction of motion corresponds to the Fibonacci bin with the largest value of $g_j$. In other words, it is the point intersected by the greatest total arc length across all great circles. The estimated direction is the winning bin: 
\begin{equation}
    \vectnorm{t}^{\pm} = \mathcal{S}[{\mathrm{argmax}(\mathcal{G}})], 
\end{equation}
where $\vectnorm{t}^{\pm}$ represents the two solutions given the symmetric nature of great circles (i.e. two great circles intersect at two locations of which both are possible solutions). The tie is resolved by solving \cref{eq:translation_set} for both solutions; details appear in the supplementary material.

\subsection{Hierarchical improvement}
The primary computational cost comes from \cref{eq:finding_theta}, where each great circle is compared to each Fibonacci bin. This produces a matrix of $\mathbf{B} \in \mathbb{R}^{N \times M}$ where each element $b_{i, j} \in \mathbf{B}$ represents the vote contribution from great circle $i$ to Fibonacci bin $\vect{s}_j$. In practice, $\mathbf{B}$ is sparse since most great circles are farther than radius $r$ from most Fibonacci bins and therefore contribute no weight. To reduce unnecessary comparisons, we use a hierarchical approach. First, we find the distance for each great circle to a sparse Fibonacci lattice with $M=1,000$ bins and radius $r_l=0.2$ to determine an approximate direction. Next, we densely sample Fibonacci bins around the top candidate region with $M=64k$ bins and radius $r_s=0.009$ to refine the direction estimate.

\subsection{Non-Linear Refinements}
Let $\mathcal{N} = \{\vect{n}_1, \ldots, \vect{n}_k\}$ be the set of normals to the great circles intersecting  the winning bin. The goal is to find a vector $\vect{p}$ that is as “orthogonal as possible” to each great circle normal $\vect{n}_i \in \mathcal{N}$. Mathematically, we want to
\begin{align}
  \text{minimize:} \quad \sum_{i} \bigl(\vect{n}_i \cdot \vect{p}\bigr)^2 \quad \text{subject to:} \quad \|\vect{p}\| = 1.
\end{align}
By rewriting the sum of squared dot products in matrix form, we get:

\begin{align}
    \sum_{i} \bigl(\vect{n}_i \cdot \vect{p}\bigr)^2  &= \sum_{i}{\vect{p}^T\vect{n}_i^T\vect{n}_i\vect{p}} \\
                                        &= \vect{p}^T\left(\sum_{i}{\vect{n}_i^T\vect{n}_i}\right)\vect{p} \\
                                        &= \vect{p}^T \mathbf{A} \vect{p},
\end{align}
where $\mathbf{A} \in \mathbb{R}^{3\times3}$ is the sum of the outer products $\vect{n}_i^T\vect{n}_i$.
When we impose the constraint $\|\vect{p}\| = 1$, this minimization problem is solved by the eigenvector of $\mathbf{A}$ corresponding to its smallest eigenvalue. We show in the experiments section that such a refinement makes a noticeable improvement to the accuracy with only a modest increase in computation.

\subsection{Early Stopping}
It is often possible to obtain a reliable heading estimate without processing every feature pair. To achieve this, we implement an early stopping criterion. We first select a random subset of $N=64$ and compute the initial estimate $\vectnorm{t}_1$. Next, we sample an additional set of $N$ features and calculate a second estimate $\vectnorm{t}_2$. If $\vectnorm{t}_1 = \vectnorm{t}_2$, the algorithm assumes convergence to the final estimate $\vectnorm{t}$; otherwise, we continue sampling $N$ additional features until convergence. Moreover, to ensure that the sampling sufficiently represents the entire space, the number of features contributing to the dominant bin must be at least $5\%$ of the total number of feature pairs.

\section{Implementation}\label{sec:implementation}

\subsection{Pre-computing arc length}
The base $b_{i, j}$ for a given great circle with normal $\vect{n}_i$ and a Fibonacci bin $\vect{s}_j$ is limited to range $0 \leq b_{i, j} \leq r$. Knowing the range allows us to subdivide the region $r$ into $k$ sections and precompute \cref{eq:lookup}. At inference time, we find the nearest precomputed base using a lookup table, which significantly improves the efficiency of our method.

\subsection{Setting the radius}\label{subsec:setting_r}
The area of each bin is given by $A = 4\pi/M$, where $M$ is the number of Fibonacci bins. This holds since the Fibonacci lattice is approximately uniformly distributed. Solving for a radius $r$ that creates a circle with area $A$, we get $r=2/\sqrt{M}$. However, empirical evaluations show that using a circle with radius $r$ as defined leaves gaps on the unit sphere. Consequently, adjusting the radius by a scaling factor $r \approx 1.15 \cdot 2/\sqrt{M}$ is advantageous to ensure better coverage of the unit sphere.

\section{Experiments}\label{sec:experiments}
\paragraph{Datasets:} We evaluate our proposed method FLIGHT for frame-to-frame direction estimation on three datasets: KITTI \cite{kitti}, TUM RGB-D \cite{tum}, and Sintel \cite{sintel}. KITTI is an outdoor autonomous driving dataset with few moving objects and relatively large baselines. Next, we evaluate on TUM RGB-D sequences as given in FM-Bench \cite{bian2019bench}. This dataset is captured indoors without moving objects. Finally, we evaluate on Sintel, a synthetic dataset with dynamic environments. It's important to evaluate on dynamic scenes to ensure robustness to outliers. 

For optical flow, we employ RapidFlow (6 iters) \cite{rapidflow}, trained on Sintel, as implemented in PTLFlow \cite{morimitsu2021ptlflow} on full-resolution frames. For the discrete correspondences, we computed both SURF \cite{surf} and SIFT \cite{SIFT} features as provided by OpenCV \cite{opencv} and use a brute-force matcher with 2 nearest neighbors. Unless stated otherwise, we use KITTI sequence $06$ for robustness \cref{subsec:robustness} and ablation \cref{subsec:ablation} studies.

\paragraph{Baselines:}  We evaluate our proposed method, FLIGHT, against four established baselines: \citet{Heikkonen1995Recovering3M} (denoted as FOE), \citet{Fredriksson2015PracticalRT} (denoted as BNB), \citet{PassiveNU} (denoted as PN), the 2-Point algorithm (see the supplementary material for further details), and on MAGSAC++ \cite{barath2019magsacfastreliableaccurate} after rotation compensation. In addition, we develop a modified variant of \cite{PassiveNU}, referred to as PN*. This variant operates on a subset of correspondences (k=200) and employs a simplified RANSAC procedure that iteratively samples until the maximum number of inliers is obtained, determined by the angular consistency between the correspondences and the estimated direction.

Given that none of the baseline implementations were publicly available, we re-implemented all methods in C++ and ran all experiments on an AMD EPYC 7543 CPU with 16 cores and 32GB of memory. For FLIGHT, we used the hierarchical approach with a sparse Fibonacci lattice of 1,000 bins and $r_l=0.2$ and a dense Fibonacci lattice with $64k$ bins and $r_s=0.009$, which we selected based on the ablation study (\cref{subsec:ablation}). Additionally, we implemented our early stopping (ES) criteria and added our non-linear refinement (NLR). To remain consistent for all experiments, we use an off-the-shelf rotation estimator \cite{PoseLib}.

\paragraph{Evaluation metrics:} We evaluate the accuracy of our method by computing the mean Average Accuracy (mAA) of the translation direction error. The translation direction error is the angle between the estimated and ground truth translation directions. Mean Average Accuracy is a standard metric in camera motion estimation and is defined as the area under the curve of the cumulative distribution function. In addition to the mAA, we also report the runtime of each method.

\subsection{Results}

\begin{table}[t!]
\centering
\resizebox{1\linewidth}{!}{\setlength{\tabcolsep}{15pt}
\begin{NiceTabular}{@{}clccc}[colortbl-like]
\toprule
\rowcolor{Gray!20}   & {\bf Method} & {\bf mAA 2°} $\uparrow$ & {\bf mAA 5°} $\uparrow$ & {\bf Time (ms)} $\downarrow$ \\
\midrule
\multirow{7}{*}{\rotatebox{90}{Flow}} 
 & FLIGHT                                & \bf{0.6193} & \bf{0.8223} & 0.9472 \\
 & FOE \cite{Heikkonen1995Recovering3M}  & 0.6065      & 0.8169      & 3.8278 \\
 & BNB \cite{Fredriksson2015PracticalRT} & 0.6013      & 0.81        & 35.7238 \\
 & PN  \cite{PassiveNU}                  & 0.5387      & 0.7796      & \bf{0.0361} \\
 & PN*                                   & 0.5754      & 0.7986      & \uf{0.8125} \\
 & 2-Points                              & \uf{0.6141} & \uf{0.8177} & 12.7976 \\
 & MAGSAC++                              & 0.6105      & 0.818       & 2.8462 \\
\midrule
\multirow{7}{*}{\rotatebox{90}{SIFT}} 
 & FLIGHT                                & 0.3929      & 0.6397      & \uf{0.5174} \\
 & FOE \cite{Heikkonen1995Recovering3M}  & \uf{0.3935} & \uf{0.6436} & 1.5053 \\
 & BNB \cite{Fredriksson2015PracticalRT} & \bf{0.4209} & \bf{0.6621} & 5.8465 \\
 & PN  \cite{PassiveNU}                  & 0.008       & 0.044       & \bf{0.0398} \\
 & PN*                                   & 0.0162      & 0.0615      & 0.6317 \\
 & 2-Points                              & 0.2715      & 0.4785      & 1.6784 \\
 & MAGSAC++                              & 0.3143      & 0.561       & 1.5754 \\
\midrule
\multirow{7}{*}{\rotatebox{90}{SURF}} 
 & FLIGHT                                & \uf{0.5548} & \bf{0.785} & 1.3762 \\
 & FOE \cite{Heikkonen1995Recovering3M}  & 0.5447      & 0.7803     & 12.3282 \\
 & BNB \cite{Fredriksson2015PracticalRT} & 0.5457      & 0.7731     & 35.0909 \\
 & PN  \cite{PassiveNU}                  & 0.0168      & 0.0847     & \bf{0.0447} \\
 & PN*                                   & 0.0684      & 0.1991     & \uf{1.1064} \\
 & 2-Points                              & \bf{0.5644} & \uf{0.7849}& 7.1115 \\
 & MAGSAC++                              & 0.4399      & 0.7091     & 4.4032 \\
\bottomrule
\end{NiceTabular}
}
\caption{\textbf{Quantitative results on KITTI dataset}. We compare the mAA of translation error and the run time of our method with multiple other baselines. Bold text indicates the best and underlines as the second best. For each method, we report the mAA at 2° and mAA 5° and Time (ms) when using Flow, SIFT, and SURF.}
\label{tab:kitti_results}
\end{table}

\paragraph{KITTI:} The results on KITTI illustrate that our method outperforms the baselines on computation time and accuracy. When using optical flow, we improve over the 2-Point baseline on mAA at $\ds{2}$ and mAA at $\ds{5}$ while running more than $92\%$ faster. Similarly, compared to FOE, we improve on mAA at $\ds{2}$ and mAA at $\ds{5}$ while running $75\%$ faster. We see even larger improvements for the other methods. \Cref{tab:kitti_results} reports the numerical results. Additionally, the results indicate that optical flow provides a more accurate estimate than SURF and SIFT across all methods. 

\begin{table}[t!]
\centering
\resizebox{1\linewidth}{!}{\setlength{\tabcolsep}{15pt}
\begin{NiceTabular}{@{}clccc}[colortbl-like]
\toprule
\rowcolor{Gray!20}   & {\bf Method} & {\bf mAA 5°} $\uparrow$ & {\bf mAA 10°} $\uparrow$ & {\bf Time (ms)} $\downarrow$ \\
\midrule
\multirow{7}{*}{\rotatebox{90}{Flow}} 
 & FLIGHT                                & 0.2731      & 0.4781      & 0.9149 \\
 & FOE \cite{Heikkonen1995Recovering3M}  & 0.2715      &  0.4795     & 43.4998 \\
 & BNB \cite{Fredriksson2015PracticalRT} & 0.1578      & 0.2988      & 29.4898 \\
 & PN  \cite{PassiveNU}                  & 0.1139      & 0.2351      & \bf{0.0375} \\
 & PN*                                   & 0.1284      & 0.2591      & \uf{0.8601} \\
 & 2-Points                              & \bf{0.2794} & \bf{0.4819} & 11.4159 \\
 & MAGSAC++                              & \uf{0.2778} & \uf{0.4827} & 3.9321 \\
\midrule
\multirow{7}{*}{\rotatebox{90}{SIFT}} 
 & FLIGHT                                & \bf{0.1102} & \bf{0.2207} & \uf{0.5116} \\
 & FOE \cite{Heikkonen1995Recovering3M}  & 0.0851      & 0.1753      & 2.2839 \\
 & BNB \cite{Fredriksson2015PracticalRT} & 0.0555      & 0.1315      & 8.3632 \\
 & PN  \cite{PassiveNU}                  & 0.0006      & 0.0035      & \bf{0.0383} \\
 & PN*                                   & 0.0         & 0.0023      & 0.6259 \\
 & 2-Points                              & 0.1008      & 0.1938      & 1.2765 \\
 & MAGSAC++                              & \uf{0.1043} & \uf{0.2149} & 4.3068 \\
\midrule
\multirow{7}{*}{\rotatebox{90}{SURF}} 
 & FLIGHT                               & \bf{0.1741} & \uf{0.3194} & \uf{0.7206} \\
 & FOE \cite{Heikkonen1995Recovering3M} & 0.1604      & 0.2905      & 18.7924 \\
 & BNB \cite{Fredriksson2015PracticalRT}& 0.0846      & 0.1858      & 13.399 \\
 & PN  \cite{PassiveNU}                 & 0.0039      & 0.0057      & \bf{0.0418} \\
 & PN*                                  & 0.0018      & 0.005       & 0.793 \\
 & 2-Points                             & \uf{0.169}  & 0.3111      & 1.9564 \\
 & MAGSAC++                             & 0.1671      & \bf{0.3236} & 9.7124 \\
\bottomrule
\end{NiceTabular}
}
\caption{\textbf{Quantitative results on TUM RGB-D dataset}. We compare the mAA of translation error and the run time of our method with multiple other baselines. Bold text indicates the best and underlines as the second best. For each method, we report the mAA at 5° and mAA at 10° and Time (ms) when using Flow, SIFT, and SURF.}
\label{tab:tum_results}
\end{table}

\paragraph{Tum RGB-D:} Next, we evaluate our method on the TUM RGB-D dataset. The dataset has a median of the magnitude of translation of just $1cm$, making it challenging to obtain meaningful estimates. Hence, we first subsample the frames to 5 FPS. In addition to extremely small translations, the frames in Tum RGB-D have significant rolling shutter distortion and are low resolution. As shown in \cref{tab:tum_results}, when using flow, our method is comparable to the 2-Points algorithm mAA at $\ds{5}$ and mAA at $\ds{10}$, while being more than 12 times faster. We also outperform the other baselines on accuracy and speed. Like KITTI, we see optical flow outperforms SIFT and SURF across all methods.  

\begin{table}[t!]
\centering
\resizebox{1\linewidth}{!}{\setlength{\tabcolsep}{15pt}
\begin{NiceTabular}{@{}clccc}[colortbl-like]
\toprule
\rowcolor{Gray!20}   & {\bf Method} & {\bf mAA 5°} $\uparrow$ & {\bf mAA 10°} $\uparrow$ & {\bf Time (ms)} $\downarrow$ \\
\midrule
\multirow{7}{*}{\rotatebox{90}{Flow}} & FLIGHT  & \textbf{0.3591} & \textbf{0.4541} & 1.4428 \\
 & FOE \cite{Heikkonen1995Recovering3M} & \underline{0.3482} & \underline{0.4429} & 132.2228 \\
 & BNB \cite{Fredriksson2015PracticalRT} & 0.1728 & 0.2834 & 233.4138 \\
 & PN  \cite{PassiveNU} & 0.1358 & 0.1806 & \textbf{0.042} \\
 & PN*  & 0.1632 & 0.2127 & \underline{0.8891} \\
 & 2-Points  & 0.291 & 0.385 & 115.2381 \\
 & MAGSAC++&         0.2815  &         0.3717  & 2.9324 \\
\midrule
\multirow{7}{*}{\rotatebox{90}{SIFT}} & FLIGHT  & \textbf{0.0936} & \textbf{0.1528} & \underline{0.6309} \\
 & FOE \cite{Heikkonen1995Recovering3M} & \underline{0.084} & \underline{0.1397} & 2.8347 \\
 & BNB \cite{Fredriksson2015PracticalRT} & 0.0547 & 0.1118 & 25.8938 \\
 & PN  \cite{PassiveNU} & 0.0159 & 0.0472 & \textbf{0.0247} \\
 & PN*  & 0.0181 & 0.0473 & 0.8004 \\
 & 2-Points  & 0.0622 & 0.1183 & 6.2423 \\
 & MAGSAC++&           0.063 &         0.1191  & 2.7891 \\
\midrule
\multirow{7}{*}{\rotatebox{90}{SURF}} & FLIGHT  & \textbf{0.2514} & \textbf{0.3369} & 1.107 \\
 & FOE \cite{Heikkonen1995Recovering3M} & \underline{0.2214} & \underline{0.2962} & 152.2203 \\
 & BNB \cite{Fredriksson2015PracticalRT} & 0.1251 & 0.2322 & 127.2539 \\
 & PN  \cite{PassiveNU} & 0.0021 & 0.0238 & \textbf{0.0279} \\
 & PN*  & 0.012 & 0.0441 & \underline{1.0323} \\
 & 2-Points  & 0.1574 & 0.2626 & 7.5353 \\
 & MAGSAC++&         0.0692          & 0.1446  & 7.4086 \\
\bottomrule
\end{NiceTabular}
}
\caption{\textbf{Quantitative results on Sintel dataset}. We compare the mAA of translation error and the run time of our method with multiple other baselines. Bold text indicates the best and underlines as the second best. For each method, we report the mAA 5° and  mAA at 10° and Time (ms) when using Flow, SIFT, and SURF.}
\label{tab:sintel_results}
\end{table}

\paragraph{Sintel:} The Sintel dataset is a synthetic dataset, but it presents its own set of challenges. Many of the sequences in Sintel have extremely small baselines. Some sequences also have moving objects that take up most of the scene's view and vote coherently. In addition, those moving objects are often on a featureless background. Those aspects make the translation estimation challenging. That said, for an mAA of $\ds{5}$ we improve over FOE by $3.1\%$, and for mAA of $\ds{10}$ by $2.5\%$ while being nearly $95$ times faster. We show our method's performance in \cref{tab:sintel_results}.

\subsection{SLAM Integration}
One direct application of FLIGHT is its ability to improve the SLAM pipeline. We integrate FLIGHT into the PySLAM \cite{freda2025pyslamopensourcemodularextensible} framework to enhance camera pose estimation. PySLAM initially estimates the camera's relative pose between frames using homography, followed by a nonlinear optimization step. Our modification introduces an intermediate step between the initial pose estimation and the nonlinear refinement. Specifically, after the initial pose, we apply the FLIGHT algorithm to obtain an estimated direction of motion. We then update the initial translation by scaling FLIGHT's direction estimate by the original magnitude estimate. The refined pose is then passed to the existing nonlinear optimization process to complete the absolute pose refinement stage. We ran PySLAM and PySLAM + FLIGHT on KITTI sequences 00 and 06 (provided by PySLAM) and on EuRoC V101, averaging results across 10 runs. On monocular SLAM (KITTI), FLIGHT reduces RMSE by 8\%, and on stereo SLAM (EuRoC) by nearly 2\%, while adding minimal computational overhead. The results are shown in \cref{tab:slam_flight}.

\begin{table}[t!]%
\centering%
\resizebox{1\linewidth}{!}{\setlength{\tabcolsep}{25pt}%
\begin{NiceTabular}{@{}clccc}[colortbl-like]%
\toprule
\rowcolor{Gray!20}   & {\bf Method} & {\bf RMSE} $\downarrow$ & {\bf STD} $\downarrow$ \\
\midrule
\multirow{2}{*}{\rotatebox{90}{\scriptsize KITTI}} & PySLAM          & 11.9354 & 4.2504 \\
                                       & PySLAM + FLIGHT & 10.9056 & 2.5515  \\
\midrule
\multirow{2}{*}{\rotatebox{90}{\scriptsize EuRoC}} & PySLAM          & 0.0895 & 0.0018 \\
                                            & PySLAM + FLIGHT & 0.0878 & 0.0005 \\
\bottomrule
\end{NiceTabular}
}
\caption{\textbf{Quantitative results on PySLAM}. We show an improvement in both error and STD when we use FLIGHT to refine the initial relative pose estimate in PySLAM. This improvement remains on monocular and stereo SLAM, with minimal computational overhead.}
\label{tab:slam_flight}
\end{table}

\subsection{Robustness} \label{subsec:robustness}
\begin{SCfigure}
    \centering
    \includegraphics[width=.505\linewidth]{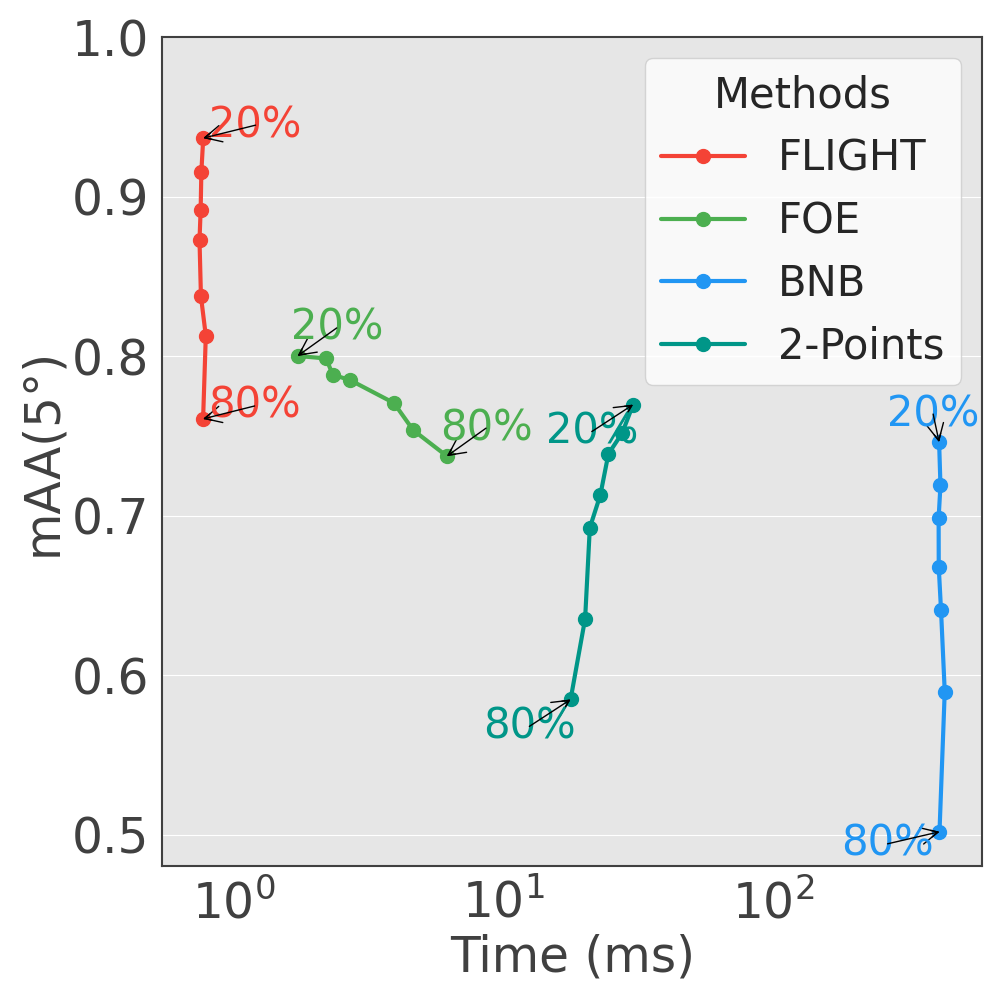}
    \caption{Results on synthetic data while varying the number of outliers. We plot time (ms) \emph{in log scale} vs accuracy for probability of outlier $p \in [20\%, \cdots ,  80\%]$ and max 2px noise. FLIGHT remains consistent in time and has higher accuracy than the baselines.}
    \label{fig:noise_outlier}
\end{SCfigure}

\paragraph{Outliers and Noise:} In the context of image matching, whether based on discrete or dense correspondence, there are two distinct types of aberrations, outliers and noise. A feature is an \emph{outlier} if, although its match is not inherently erroneous, it fails to conform to the global model. This typically occurs when the feature belongs to an independently moving object. In contrast, an aberration is considered \emph{noise} when minor errors are introduced during the feature estimation process.

We evaluate robust methods to outliers and noise on synthetically generated data. We start by sampling 500 random directions on the unit sphere. Given a random direction $\vectnorm{t} = [t_1, t_2, t_3]$, we compute the optical flow resulting from the camera translation in the direction of $\vectnorm{t}$ using the perspective projection equation:
\begin{align}
u_i = -f t_1 + x_i t_3, \qquad
v_i = -f t_2 + y_i t_3, 
\end{align}
where $[u_i, v_i]$ is the optical flow and $f=576.0$ is the focal length. To simulate the impact of outliers on both accuracy and processing time, each flow vector is assigned a probability $p \in [20\%, 30\%, 40\%, 50\%, 60\%, 70\%, 80\%]$ of being replaced with a random flow vector. Additionally, Gaussian noise is added to every flow vector, with the deviation limited to 2 pixels, regardless of whether the vector was replaced.   

\Cref{fig:noise_outlier} shows the performance of each model under varying levels of outliers. As the percentage of outliers increases, FLIGHT's runtime remains stable, which aligns with the findings presented in \cref{tab:kitti_results}, \cref{tab:tum_results}, and \cref{tab:sintel_results}. Although accuracy decreases with the addition of more outliers, FLIGHT consistently outperforms the state-of-the-art.

\begin{SCfigure}
    \centering
    \includegraphics[width=.55\linewidth]{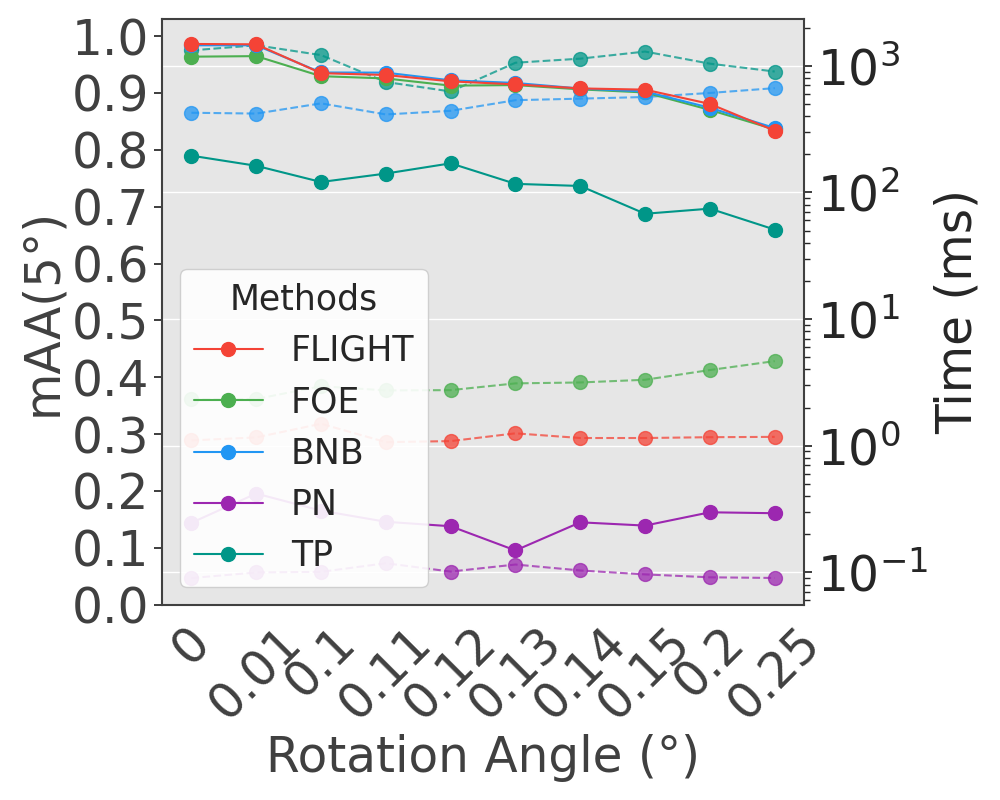}
    \caption{ The plot shows both mAA $\ds{5}$ and time (log scale) in ms under varying rotation noise. The solid line shows the mAA, and the dashed lines show the time. Up to $\ds{.15}$ rotation noise FLIGHT mAA remains above .9 while maintaining a near constant runtime.}
    \label{fig:roterr}
\end{SCfigure}
\paragraph{Rotation error:} We evaluate the effect of rotation noise on direction estimation by sampling 500 random directions on the unit sphere. Here, we inject 20\% outliers, and add Gaussian noise up to 2px. We then apply Gaussian rotation noise ranging up to $\ds{.25}$, although modern IMUs rarely exceed $\ds{.01}$ \cite{Delattre2023RobustRotation}. \Cref{fig:roterr} reports the mAA $\ds{5}$ and the runtime (ms) for each method. Under rotation errors up to $\ds{.15}$, FLIGHT maintains an mAA $\ds{5}$ of  $\approx 0.9$ or higher and a near constant runtime,  outperforming all other robust approaches. Beyond this noise level, all methods degrade at a similar rate. These results demonstrate that, within realistic rotation noise (below $\ds{.15}$), FLIGHT offers reliable improvements on accuracy and speed.

\subsection{Ablation study} \label{subsec:ablation}
In \cref{sec:method}, we described several techniques we use to improve FLIGHT's speed and accuracy. In this section, we evaluate the effects of these techniques and how we selected the parameters. 

\begin{table}[t!]
\centering
\resizebox{1\linewidth}{!}{\setlength{\tabcolsep}{8pt}
\begin{NiceTabular}{@{}clcccc}[colortbl-like]
\toprule
\rowcolor{Gray!20} {\bf Hierarchical } & {\bf NLR} & {\bf ES} & {\bf mAA 5°} $\uparrow$ & {\bf mAA 10°} $\uparrow$ & {\bf Time (ms)} $\downarrow$ \\
\midrule
                                &            &             & 0.9087             & 0.9568             & 472.2631 \\           
\rowcolor{Gray!10}              &            & \checkmark  & 0.9083             & 0.9565             & 127.6043 \\
                                & \checkmark &             & 0.9123             &  0.9586            & 472.1992 \\
\rowcolor{Gray!10}              & \checkmark & \checkmark  &  0.9122            &  0.9585            & 127.0941 \\
\checkmark                      &            &             & 0.9073             & 0.9561             & 1.2672 \\
\rowcolor{Gray!10}\checkmark    &            & \checkmark  & 0.9067             & 0.9558             & \textbf{0.8593} \\
\checkmark                      & \checkmark &             & \textbf{0.9131}    & \textbf{0.9590}    & 1.4531 \\
\rowcolor{Gray!10}\checkmark    & \checkmark &  \checkmark & \underline{0.9128} & \underline{0.9589} & \underline{0.9778} \\
\bottomrule
\end{NiceTabular}
}
\caption{Quantitative results on techniques to improve our base method to obtain our FLIGHT method. The experiments show that using the hierarchical approach with NLR and early stopping provided the best trade-off between speed and accuracy. }
\label{tab:settings_results}
\end{table}

\paragraph{Base Model Improvements:} \Cref{tab:settings_results} presents the quantitative evaluation of various configurations of our proposed method. In the base model, we sample a dense Fibonacci lattice without incorporating early stopping (ES) or non-linear refinement (NLR). Incorporating a hierarchical approach reduces the computational time by over $99\%$, and incorporating ES decreases the computational time by an additional $32\%$. We also see an improvement in accuracy by adding an NLR step, although at a slight speed cost. Overall, we achieve the optimal trade-off between computational cost and accuracy by implementing a hierarchical strategy combined with ES and NLR, which we referred to in this work as \textbf{FLIGHT}.

\begin{figure}
    \centering
     \begin{subfigure}[b]{0.23\textwidth}
         \centering
         \includegraphics[width=\textwidth]{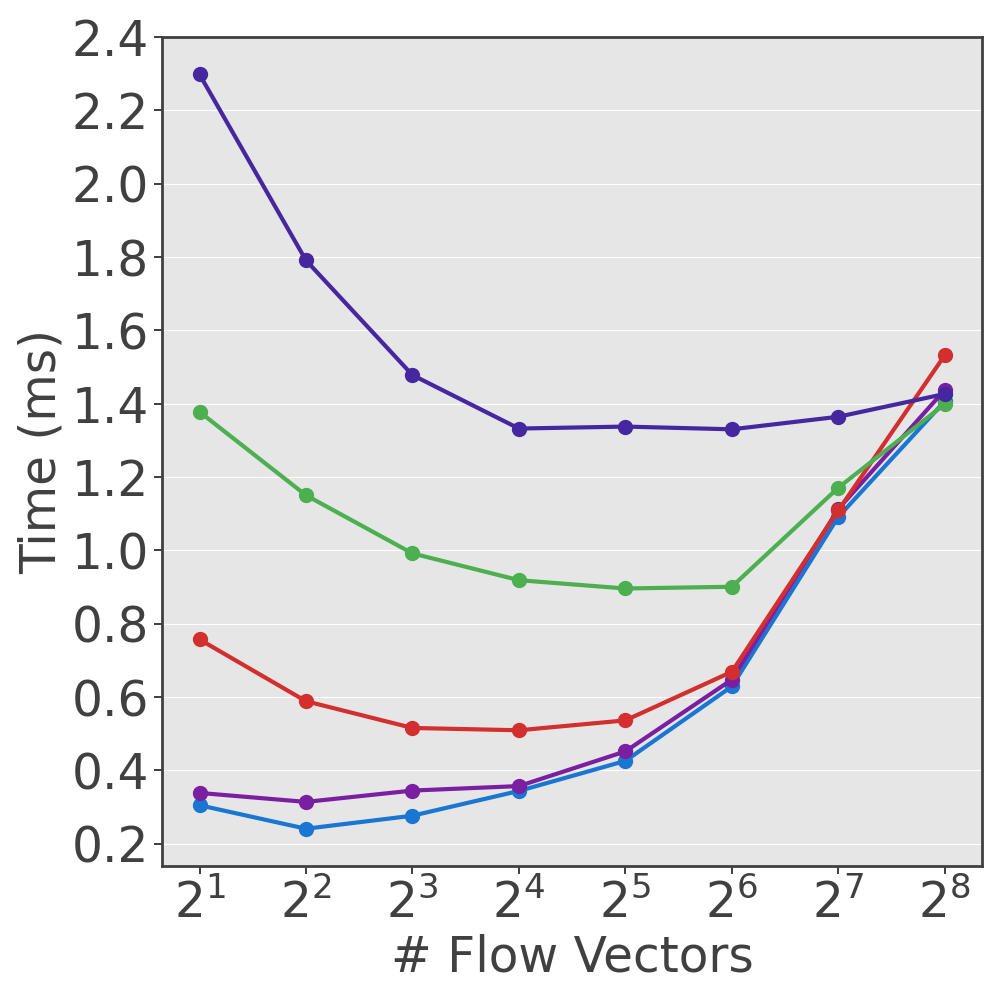}
         \label{fig:es_boxsize_vs_time}
     \end{subfigure}
     \begin{subfigure}[b]{0.225\textwidth}
         \centering
         \includegraphics[width=\textwidth]{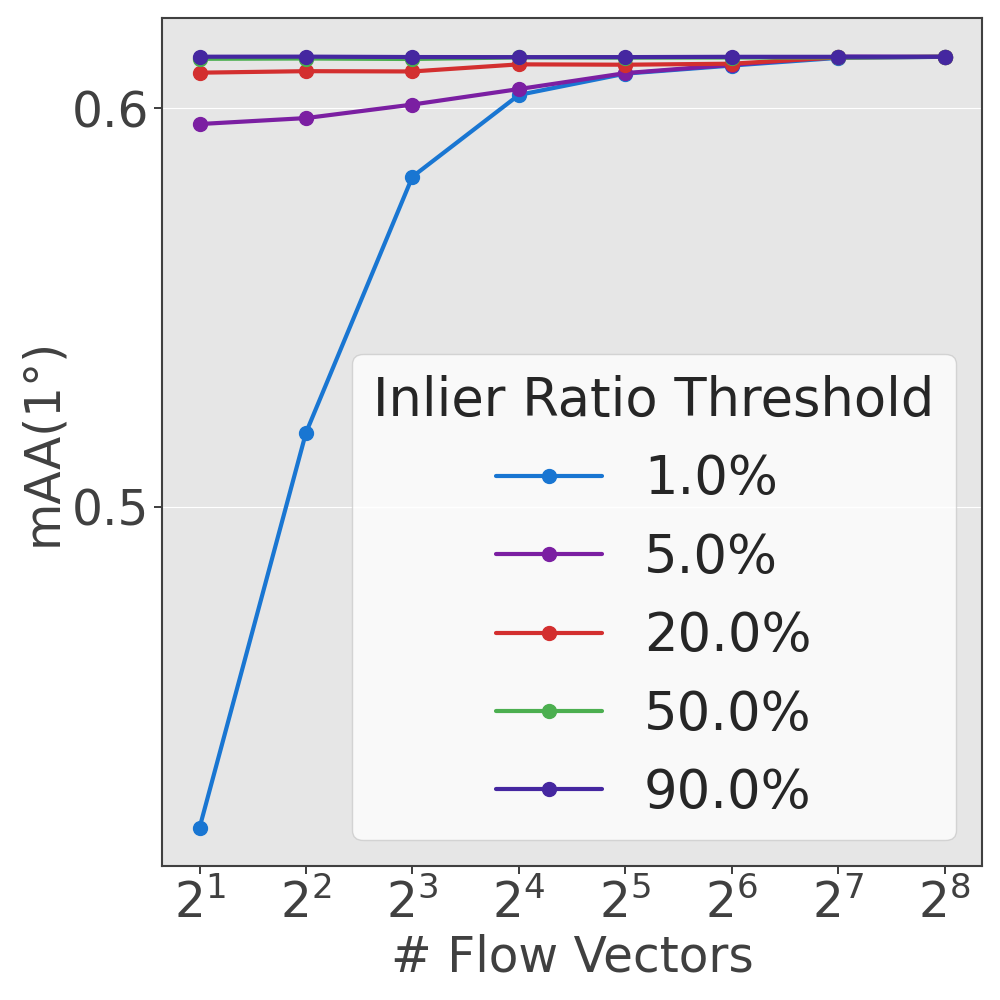}
         \label{fig:es_boxsize_vs_acc}
     \end{subfigure}
     \caption{\textbf{Left:} We compare the sample size $N$ vs. the runtime (ms) for FLIGHT for different values of $p \in [1\%, 5\%, 20\%, 50\%, 90\%]$. \textbf{Right:} We plot sample size $N$ v.s. accuracy under the same conditions. }
     \label{fig:es_settings}
\end{figure}

\paragraph{Early Stopping (ES) Criteria:} The ES approach we implemented consists of two hyperparameters: 1) the number of correspondences $N$ we sample at each step, and 2) the minimum proportion $p  \in [1\%, 5\%, 20\%, 50\%, 90\%]$ of votes required for the winning bin relative to the total number correspondences. In \cref{fig:es_settings}, we present a comparative analysis of accuracy and computational time as functions of $N$ while varying $p$. Our findings indicate that setting $N=64$ and $p=0.05$ yields an optimal balance between accuracy and efficiency. Specifically, when $N=64$, the accuracy converges to that of the optimal solution (i.e., when all correspondences are considered) with only a marginal increase in computational cost. Additionally, setting $p=0.05$ is justified since it maintains consistent accuracy (relative to lower values) without incurring significant additional runtime costs.

\begin{table}[t!]
\centering
\resizebox{1\linewidth}{!}{\setlength{\tabcolsep}{15pt}
\begin{NiceTabular}{cccc}[colortbl-like]
\toprule
\rowcolor{Gray!20} {\bf \#Bins}$\times 1000$ & {\bf mAA 1°} $\uparrow$ & {\bf mAA 5°} $\uparrow$ & {\bf Time (ms)} $\downarrow$ \\
\midrule
49 & 0.5765 & 0.9062 & \underline{0.6432} \\
54 & 0.5735 & 0.9050 & \textbf{0.6291} \\
59 & 0.6020 & 0.9107 & 0.6726 \\
64 & \textbf{0.6088} & \textbf{0.9123} & 0.6833 \\
68 & 0.5926 & 0.9090 & 0.6697 \\
102 & 0.5933 & 0.9092 & 0.7662 \\
107 & \underline{0.6047} & \underline{0.9117} & 0.7495 \\
112 & 0.5866 & 0.9079 & 0.8121 \\
\bottomrule
\end{NiceTabular}}
\caption{The number of bins (in thousands) and their corresponding effect on accuracy vs speed. Using a Fibonacci lattice of $64k$ provides a good balance of speed vs accuracy. }
\label{tab:bins_num}
\end{table}

\paragraph{Number of bins:} Selecting the Fibonacci lattice sampling (number of bins) is a balancing act between accuracy and computational cost. Sparse sampling increases the heading estimation error, while dense sampling increases the computational cost. In the non-hierarchical method, this cost is significant, whereas in the hierarchical setting, the impact is limited because \cref{eq:finding_theta} is solved only on a subset of bins. \cref{tab:bins_num} presents the observed trade-off between accuracy and runtime for a range of bin counts. In our implementation, we use $64k$ bins, which provides a practical balance between precision and efficiency. When sampling the dense Fibonacci lattice, $64k$ bins show the best accuracy on both mAA 1° and mAA 5°, while only being slightly slower.  

\section{Conclusion}
We introduce FLIGHT, a robust method for estimating camera translation direction using a generalization of the Hough transform on $\mathcal{S}^2$. Experiments on three benchmarks show that it is on the Pareto frontier of accuracy versus efficiency and runs in real time. Additionally, our methods' runtime remains stable even as the level of outliers and noise increases. Finally, we demonstrate that integrating FLIGHT into an SLAM pipeline reduces the final trajectory error. These results point to FLIGHT as a practical component for systems that require fast, reliable translation direction estimates.

{
    \small
    \bibliographystyle{ieeenat_fullname}
    \bibliography{main}
}

\end{document}